\documentclass[12pt,letterpaper,copyedit]{article}
\usepackage{osajnl2}
\usepackage{graphicx,color}
\usepackage{paralist}
\usepackage{amsmath}
\usepackage{amsfonts}
\usepackage{subfigure}
\usepackage{hyperref}
\usepackage{url}
\usepackage{breakurl}
\usepackage{algorithm2e}
\usepackage{comment}
\usepackage{fancyhdr}

\newcommand{\bx}{{\bf x}}
\newcommand{\bc}{{\bf c}}
\newcommand{\bv}{{\bf v}}

\begin{document}

\title{Fast Color Quantization Using Weighted Sort-Means Clustering}
\author{M. Emre Celebi}
\address{Dept.\ of Computer Science, Louisiana State University, Shreveport, LA, USA}
\email{ecelebi@lsus.edu}

\begin{abstract}

Color quantization is an important operation with numerous applications in graphics and image processing. Most quantization methods are essentially based on data clustering algorithms. However, despite its popularity as a general purpose clustering algorithm, k-means has not received much respect in the color quantization literature because of its high computational requirements and sensitivity to initialization. In this paper, a fast color quantization method based on k-means is presented. The method involves several modifications to the conventional (batch) k-means algorithm including data reduction, sample weighting, and the use of triangle inequality to speed up the nearest neighbor search. Experiments on a diverse set of images demonstrate that, with the proposed modifications, k-means becomes very competitive with state-of-the-art color quantization methods in terms of both effectiveness and efficiency.

\end{abstract}


\maketitle

\section{Introduction}
\label{sec_intro}

True-color images typically contain thousands of colors, which makes their display, storage, transmission, and processing problematic. For this reason, color quantization (reduction) is commonly used as a preprocessing step for various graphics and image processing tasks. In the past, color quantization was a necessity due to the limitations of the display hardware, which could not handle the 16 million possible colors in 24-bit images. Although 24-bit display hardware has become more common, color quantization still maintains its practical value \cite{Brun02}. Modern applications of color quantization include:
\begin{inparaenum}[(i)]
\item image compression \cite{Yang98},
\item image segmentation \cite{Deng01a},
\item image analysis \cite{Sertel08},
\item image watermarking \cite{Kuo07}, and
\item content-based image retrieval \cite{Deng01b}.
\end{inparaenum}
\par
The process of color quantization is mainly comprised of two phases: palette design (the selection of a small set of colors that represents the original image colors) and pixel mapping (the assignment of each input pixel to one of the palette colors). The primary objective is to reduce the number of unique colors, $N'$, in an image to $K$ ($K \ll N'$) with minimal distortion. In most applications, 24-bit pixels in the original image are reduced to 8 bits or fewer. Since natural images often contain a large number of colors, faithful representation of these images with a limited size palette is a difficult problem.
\par
Color quantization methods can be broadly classified into two categories \cite{Xiang07}: image-independent methods that determine a universal (fixed) palette without regard to any specific image \cite{Gentile90}, and image-dependent methods that determine a custom (adaptive) palette based on the color distribution of the images. Despite being very fast, image-independent methods usually give poor results since they do not take into account the image contents. Therefore, most of the studies in the literature consider only image-dependent methods, which strive to achieve a better balance between computational efficiency and visual quality of the quantization output.
\par
Numerous image-dependent color quantization methods have been developed in the past three decades. These can be categorized into two  families: preclustering methods and postclustering methods \cite{Brun02}. Preclustering methods are mostly based on the statistical analysis of the color distribution of the images. Divisive preclustering methods start with a single cluster that contains all $N$ image pixels. This initial cluster is recursively subdivided until $K$ clusters are obtained. Well-known divisive methods include median-cut \cite{Heckbert82}, octree \cite{Gervautz88}, variance-based method \cite{Wan90}, binary splitting \cite{Orchard91}, greedy orthogonal bipartitioning \cite{Wu91}, center-cut \cite{Joy93}, and rwm-cut \cite{Yang96}. More recent methods can be found in \cite{Cheng01,Sirisathitkul04,Kanjanawanishkul05}. On the other hand, agglomerative preclustering methods \cite{Equitz89,Balasubramanian91,Xiang94,Velho97,Brun00} start with $N$ singleton clusters each of which contains one image pixel. These clusters are repeatedly merged until $K$ clusters remain. In contrast to preclustering methods that compute the palette only once, postclutering methods first determine an initial palette and then improve it iteratively. Essentially, any data clustering method can be used for this purpose. Since these methods involve iterative or stochastic optimization, they can obtain higher quality results when compared to preclustering methods at the expense of increased computational time. Clustering algorithms adapted to color quantization include k-means \cite{Kasuga00,Huang04,Hu07,Hu08}, minmax \cite{Xiang97}, competitive learning \cite{Verevka95,Scheunders97,Celebi09}, fuzzy c-means \cite{Ozdemir02,Schaefer09}, BIRCH \cite{Bing04}, and self-organizing maps \cite{Dekker94,Papamarkos02,Chang05}.
\par
In this paper, a fast color quantization method based on the k-means clustering algorithm \cite{Linde80} is presented. The method first reduces the amount of data to be clustered by sampling only the pixels with unique colors. In order to incorporate the color distribution of the pixels into the clustering procedure, each color sample is assigned a weight proportional to its frequency. These weighted samples are then clustered using a fast and exact variant of the k-means algorithm. The set of final cluster centers is taken as the quantization palette.
\par
The rest of the paper is organized as follows. Section \ref{sec_kmeans} describes the conventional k-means clustering algorithm and the proposed modifications. Section \ref{sec_exp} describes the experimental setup and presents the comparison of the proposed method with other color quantization methods. Finally, Section \ref{sec_conc} gives the conclusions.

\section{Color Quantization Using K-Means Clustering Algorithm}
\label{sec_kmeans}

The k-means (KM) algorithm is inarguably one of the most widely used methods for data clustering \cite{Gan07}. Given a data set $X = \left\{ \bx_1, \ldots, \bx_N \right\} \in \mathbb{R}^D $, the objective of KM is to partition $X$ into $K$ exhaustive and mutually exclusive clusters $S = \left\{ S_1, \ldots, S_k \right\},\;\; \bigcup\nolimits_{k = 1}^K {S_k = X},\;\; S_i \cap S_j \equiv \emptyset$ for $i \neq j$ by minimizing the sum of squared error (SSE):

\begin{equation}
\label{eq_sse}
 \mbox{SSE} = \sum\limits_{k = 1}^K {\sum\limits_{\bx_i \in S_k} {\left\| {\bx_i - \bc_k } \right\|_2^2 } }
\end{equation}

where, $\| \, \|_2$ denotes the Euclidean ($L_2$) norm and $\bc_k$ is the center of cluster $S_k$ calculated as the mean of the points that belong to this cluster. This problem is known to be computationally intractable even for $K = 2$ \cite{Drineas04}, but a heuristic method developed by Lloyd \cite{Lloyd82} offers a simple solution. Lloyd's algorithm starts with $K$ arbitrary centers, typically chosen uniformly at random from the data points \cite{Forgy65}. Each point is then assigned to the nearest center, and each center is recalculated as the mean of all points assigned to it. These two steps are repeated until a predefined termination criterion is met. The pseudocode for this procedure is given in Algo.\ (\ref{algo_kmeans}) (\textbf{bold} symbols denote vectors). Here, $m[i]$ denotes the membership of point $\bx_i$, i.e.\ index of the cluster center that is nearest to $\bx_i$.

\begin{algorithm}
\linespread{1}
\normalsize
{
\label{algo_kmeans}
\SetKwInOut{Input}{input}
\SetKwInOut{Output}{output}
\Input { $ X = \left\{ \bx_1, \ldots, \bx_N \right\} \in \mathbb{R}^D $ ($N \times D$ input data set) }
\Output { $ C = \left\{ \bc_1, \ldots, \bc_K \right\} \in \mathbb{R}^D $ ($K$ cluster centers)}
Select a random subset $C$ of $X$ as the initial set of cluster centers\;
\While { termination criterion is not met }
{
 \For{$( i = 1; i \leq N; i = i + 1 )$}
  {
   \CommentSty{Assign $\bx_i$ to the nearest cluster}\;
   $m[i] = \underset{k \in \left\{ 1, \ldots, K \right\} }{\operatorname{argmin}} \left\| \bx_i - \bc_k \right\|^2$\;
  }
 \CommentSty{Recalculate the cluster centers}\;
 \For{$( k = 1; k \leq K; k = k + 1 )$}
  {
   \CommentSty{Cluster $S_k$ contains the set of points $\bx_i$ that are nearest to the center $\bc_k$}\;
   $S_k  = \left\{ {\bx_i \left| {m[i] = k} \right.} \right\}$\;
   \CommentSty{Calculate the new center $\bc_k$ as the mean of the points that belong to $S_k$}\;
   $\bc_k  = \frac{1} {{\left| {S_k } \right|}}\sum\limits_{\bx_i \in S_k } {\bx_i }$\;
  }
}
\caption{Conventional K-Means Algorithm}
}
\end{algorithm}

When compared to the preclustering methods, there are two problems with using KM for color quantization. First, due to its iterative nature, the algorithm might require an excessive amount of time to obtain an acceptable output quality. Second, the output is quite sensitive to the initial choice of the cluster centers. In order to address these problems, we propose several modifications to the conventional KM algorithm:

\begin{itemize}
 \item \textbf{Data sampling}: A straightforward way to speed up KM is to reduce the amount of data, which can be achieved by sampling the original image. Although random sampling can be used for this purpose, there are two problems with this approach. First, random sampling will further destabilize the clustering procedure in the sense that the output will be less predictable. Second, sampling rate will be an additional parameter that will have a significant impact on the output. In order to avoid these drawbacks, we propose a deterministic sampling strategy in which only the pixels with unique colors are sampled. The unique colors in an image can be determined efficiently using a hash table that uses chaining for collision resolution and a universal hash function of the form: $h_a(\bx) = \left( {\sum\nolimits_{i = 1}^3 {a_i x_i}} \right)\bmod m$, where $\bx = (x_1, x_2, x_3)$ denotes a pixel with red ($x_1$), green ($x_2$), and blue ($x_3$) components, $m$ is a prime number, and the elements of sequence $a = (a_1, a_2, a_3)$ are chosen randomly from the set $\left\{ 0, 1, \ldots, m - 1\right\}$.

 \item \textbf{Sample weighting}: An important disadvantage of the proposed sampling strategy is that it disregards the color distribution of the original image. In order to address this problem, each point is assigned a weight that is proportional to its frequency (note that the frequency information is collected during the data sampling stage). The weights are normalized by the number of pixels in the image to avoid numerical instabilities in the calculations. In addition, Algo.\ (\ref{algo_kmeans}) is modified to incorporate the weights in the clustering procedure.

 \item \textbf{Sort-Means algorithm}: The assignment phase of KM involves many redundant distance calculations. In particular, for each point, the distances to each of the $K$ cluster centers are calculated. Consider a point $\bx_i$, two cluster centers $\bc_a$ and $\bc_b$ and a distance metric $d$, using the triangle inequality, we have $ d(\bc_a,\bc_b) \leq d(\bx_i,\bc_a) + d(\bx_i,\bc_b) $. Therefore, if we know that $ 2d(\bx_i,\bc_a) \leq d(\bc_a,\bc_b) $, we can conclude that $ d(\bx_i,\bc_a) \leq d(\bx_i,\bc_b) $ without having to calculate $ d(\bx_i,\bc_b) $. The compare-means algorithm \cite{Phillips02} precalculates the pairwise distances between cluster centers at the beginning of each iteration. When searching for the nearest cluster center for each point, the algorithm often avoids a large number of distance calculations with the help of the triangle inequality test. The sort-means (SM) algorithm \cite{Phillips02} further reduces the number of distance calculations by sorting the distance values associated with each cluster center in ascending order. At each iteration, point $\bx_i$ is compared against the cluster centers in increasing order of distance from the center $\bc_k$ that $\bx_i$ was assigned to in the previous iteration. If a center that is far enough from $\bc_k$ is reached, all of the remaining centers can be skipped and the procedure continues with the next point. In this way, SM avoids the overhead of going through all the centers. It should be noted that more elaborate approaches to accelerate KM have been proposed in the literature. These include algorithms based on kd-trees \cite{Kanungo02}, coresets \cite{Har-Peled04}, and more sophisticated uses of the triangle inequality \cite{Elkan03}. Some of these algorithms \cite{Har-Peled04,Elkan03} are not suitable for low dimensional data sets such as color image data since they incur significant overhead to create and update auxiliary data structures \cite{Elkan03}. Others \cite{Kanungo02} provide computational gains comparable to SM at the expense of significant conceptual and implementation complexity. In contrast, SM is conceptually simple, easy to implement, and incurs very small overhead, which makes it an ideal candidate for color clustering.
\end{itemize}

We refer to the KM algorithm with the abovementioned modifications as the 'Weighted Sort-Means' (WSM) algorithm. The pseudocode for WSM is given in Algo.\ (\ref{algo_wsmeans}).

\begin{algorithm}
\linespread{1}
\normalsize
{
\label{algo_wsmeans}
\SetKwInOut{Input}{input}
\SetKwInOut{Output}{output}
\Input { $ X = \left\{ \bx_1, \ldots, \bx_{N'} \right\} \in \mathbb{R}^D $ ($N' \times D$ input data set)\\
         $ W = \left\{ w_1, \ldots, w_{N'} \right\} \in [0,1] $ ($N'$ point weights) }
\Output { $ C = \left\{ \bc_1, \ldots, \bc_K \right\} \in \mathbb{R}^D $ ($K$ cluster centers) }
Select a random subset $C$ of $X$ as the initial set of cluster centers\;
\While { termination criterion is not met }
{
 \CommentSty{Calculate the pairwise distances between the cluster centers}\;
 \For{$( i = 1; i \leq K; i = i + 1 )$}
  {
   \For{$( j = i + 1; j \leq K; j = j + 1 )$}
    {
     $d[i][j] = d[j][i] = \| \bc_i - \bc_j \|^2$\;
    }
  }

 \CommentSty{Construct a $K \times K$ matrix $M$ in which row $i$ is a permutation of $1, \ldots K$ that
 represents the clusters in increasing order of distance of their centers from $\bc_i$}\;

 \For{$( i = 1; i \leq N'; i = i + 1 )$}
  {
   \CommentSty{Let $S_p$ be the cluster that $\bx_i$ was assigned to in the previous iteration}\;
   $p = m[i]$\;
   min\_dist $=$ prev\_dist $= \| \bx_i - \bc_p \|^2$\;

   \CommentSty{Update the nearest center if necessary}\;
   \For{$( j = 2; j \leq K; j = j + 1 )$}
    {
     $t = M[p][j]$\;
     \If{$d[p][t] \geq 4 \; prev\_dist$}
      {
       \CommentSty{There can be no other closer center. Stop checking}\;
       break\;
      }

     dist = $\| \bx_i - \bc_t \|^2$\;

     \If{$dist \leq min\_dist$}
      {
       \CommentSty{$\bc_t$ is closer to $\bx_i$ than $\bc_p$}\;
       min\_dist = dist\;
       $m[i] = t$;
      }
    }
  }
 \CommentSty{Recalculate the cluster centers}\;
 \For{$( k = 1; k \leq K; k = k + 1 )$}
  {
   \CommentSty{Calculate the new center $\bc_k$ as the weighted mean\\
    of points that are nearest to it}\;
   $\bc_k = {{\left( {\sum\limits_{m[i] = k} {w_i \bx_i } } \right)} \mathord{\left/
   {\vphantom {{\left( {\sum\limits_{m[i] = k} {w_i \bx_i } } \right)} {\sum\limits_{m[i] = k} {w_i } }}} \right.
   \kern-\nulldelimiterspace} {\sum\limits_{m[i] = k} {w_i } }}$\;
  }
}
\caption{Weighted Sort-Means Algorithm}
}
\end{algorithm}

\section{Experimental Results and Discussion}
\label{sec_exp}

\subsection{Image set and performance criteria}
\label{sec_criteria}

The proposed method was tested on some of the most commonly used test images in the quantization literature. The natural images in the set included Airplane ($512 \times 512$, 77,041 (29\%) unique colors), Baboon ($512 \times 512$, 153,171 (58\%) unique colors), Boats ($787 \times 576$, 140,971 (31\%) unique colors), Lenna ($512 \times 480$, 56,164 (23\%) unique colors), Parrots ($1536 \times 1024$, 200,611 (13\%) unique colors), and Peppers ($512 \times 512$, 111,344 (42\%) unique colors). The synthetic images included Fish ($300 \times 200$, 28,170 (47\%) unique colors) and Poolballs ($510 \times 383$, 13,604 (7\%) unique colors).

The effectiveness of a quantization method was quantified by the Mean Squared Error (MSE) measure:
\begin{equation}
 \mbox{MSE}\left( {{\bf X},{\bf \hat{X}}} \right) = \frac{1}{{HW}}\sum\nolimits_{h = 1}^H {\sum\nolimits_{w = 1}^W {\parallel {\bf x}(h,w) - {\bf \hat{x}}(h,w) \parallel}_2^2}
\end{equation}
where ${\bf X}$ and ${\bf \hat{X}}$ denote respectively the $H \times W$ original and quantized images in the RGB color space. MSE represents the average distortion with respect to the $L_2^2$ norm \eqref{eq_sse} and is the most commonly used evaluation measure in the quantization literature \cite{Brun02,Xiang07}. Note that the Peak Signal-to-Noise Ratio (PSNR) measure can be easily calculated from the MSE value:
\begin{equation}
 \mbox{PSNR} = 20 \log_{10} \left( {\frac{{255}}{{\sqrt {\mbox{MSE}} }}} \right).
\end{equation}

The efficiency of a quantization method was measured by CPU time in milliseconds. Note that only the palette generation phase was considered since this is the most time consuming part of the majority of quantization methods. All of the programs were implemented in the C language, compiled with the gcc v4.2.4 compiler, and executed on an Intel\textregistered Core\texttrademark2 Quad Q6700 2.66GHz machine. The time figures were averaged over 100 runs.

\subsection{Comparison of WSM against other quantization methods}
\label{sec_comp}

The WSM algorithm was compared to some of the well-known quantization methods in the literature:

\begin{itemize}
 \item \textbf{Median-cut (MC)} \cite{Heckbert82}: This method starts by building a $32 \times 32 \times 32$ color histogram that contains the original pixel values reduced to 5 bits per channel by uniform quantization. This histogram volume is then recursively split into smaller boxes until $K$ boxes are obtained. At each step, the box that contains the largest number of pixels is split along the longest axis at the median point, so that the resulting subboxes each contain approximately the same number of pixels. The centroids of the final $K$ boxes are taken as the color palette.

 \item \textbf{Variance-based method (WAN)} \cite{Wan90}: This method is similar to MC, with the exception that at each step the box with the largest weighted variance (squared error) is split along the major (principal) axis at the point that minimizes the marginal squared error.

 \item \textbf{Greedy orthogonal bipartitioning (WU)} \cite{Wu91}: This method is similar to WAN, with the exception that at each step the box with the largest weighted variance is split along the axis that minimizes the sum of the variances on both sides.

 \item \textbf{Neu-quant (NEU)} \cite{Dekker94}: This method utilizes a one-dimensional self-organizing map (Kohonen neural network) with 256 neurons. A random subset of $N/f$ pixels is used in the training phase and the final weights of the neurons are taken as the color palette. In the experiments, the highest quality configuration, i.e.\ $f = 1$, was used.

 \item \textbf{Modified minmax (MMM)} \cite{Xiang97}: This method choses the first center $\bc_1$ arbitrarily from the data set and the $i$-th center $\bc_i$ ($i = 2, \ldots, K$) is chosen to be the point that has the largest minimum weighted $L_2^2$ distance (the weights for the red, green, and blue channels are taken as 0.5, 1.0, and 0.25, respectively) to the previously selected centers, i.e.\ $\bc_1, \bc_2, \ldots, \bc_{i-1}$. Each of these initial centers is then recalculated as the mean of the points assigned to it.

 \item \textbf{Split \& Merge (SAM)} \cite{Brun00}: This two-phase method first divides the color space uniformly into $B$ partitions. This initial set of $B$ clusters is represented as an adjacency graph. In the second phase, $(B - K)$ merge operations are performed to obtain the final $K$ clusters. At each step of the second phase, the pair of clusters with the minimum joint quantization error are merged. In the experiments, the initial number of clusters was set to $B = 20 K$.

 \item \textbf{Fuzzy c-means (FCM)} \cite{Bezdek81}: FCM is a generalization of KM in which points can belong to more than one cluster. The algorithm involves the minimization of the functional $J_q(U,V) = \sum\nolimits_{i = 1}^N {\sum\nolimits_{k = 1}^K {u_{ik}^q \left\| {\bx_i - \bv_k } \right\|_2^2 } }$ with respect to $U$ (a fuzzy $K$-partition of the data set) and $V$ (a set of prototypes -- cluster centers). The parameter $q$ controls the fuzziness of the resulting clusters. At each iteration, the membership matrix $U$ is updated by
 $u_{ik}  = \left( {\sum\nolimits_{j = 1}^K {\left( {{{\left\| {\bx_i - \bv_k } \right\|_2 } \mathord{\left/
 {\vphantom {{\left\| {\bx_i - \bv_k } \right\|_2 } {\left\| {\bx_i - \bv_j } \right\|_2 }}} \right.
 \kern-\nulldelimiterspace} {\left\| {\bx_i - \bv_j } \right\|_2 }}} \right)^{{2 \mathord{\left/
 {\vphantom {2 {(q - 1)}}} \right. \kern-\nulldelimiterspace} {(q - 1)}}} } } \right)^{ - 1}$,
which is followed by the update of the prototype matrix $V$ by
$\bv_k  = {{\left( {\sum\nolimits_{i = 1}^N {u^q _{ik} \bx_i } } \right)} \mathord{\left/ {\vphantom {{\left( {\sum\nolimits_{i = 1}^N {u^q _{ik} \bx_i } } \right)} {\left( {\sum\nolimits_{i = 1}^N {u^q _{ik} } } \right)}}} \right. \kern-\nulldelimiterspace} {\left( {\sum\nolimits_{i = 1}^N {u^q _{ik} } } \right)}}$.
A n\"{a}ive implementation of the FCM algorithm has a complexity that is quadratic in $K$. In the experiments, a linear complexity formulation described in \cite{Kolen02} was used and the fuzziness parameter was set to $q=2$ as commonly seen in the fuzzy clustering literature \cite{Gan07}.

 \item \textbf{Fuzzy c-means with partition index maximization (PIM)} \cite{Ozdemir02}: This method is an extension of FCM in which the functional to be minimized incorporates a cluster validity measure called the 'partition index' (PI). This index measures how well a point $\bx_i$ has been classified and is defined as $P_i  = \sum\nolimits_{k = 1}^K {u^q _{ik} }$. The FCM functional can be modified to incorporate PI as follows: $J^\alpha_q(U,V) = \sum\nolimits_{i = 1}^N {\sum\nolimits_{k = 1}^K {u_{ik}^q \left\| {\bx_i - \bv_k } \right\|_2^2 } } - \alpha \sum\nolimits_{i = 1}^N {P_i}$. The parameter $\alpha$ controls the weight of the second term. The procedure that minimizes $J^\alpha_q(U,V)$ is identical to the one used in FCM except for the membership matrix update equation:
 $u_{ik}  = \left( {\sum\nolimits_{j = 1}^K {\left[ {{{\left( {\left\| {\bx_i  - \bv_k } \right\|_2  - \alpha } \right)} \mathord{\left/
 {\vphantom {{\left( {\left\| {\bx_i  - \bv_k } \right\|_2  - \alpha } \right)} {\left( {\left\| {\bx_i - \bv_j } \right\|_2  - \alpha } \right)}}} \right. \kern-\nulldelimiterspace} {\left( {\left\| {\bx_i - \bv_j } \right\|_2  - \alpha} \right)}}} \right]^{{2 \mathord{\left/
 {\vphantom {2 {(q - 1)}}} \right. \kern-\nulldelimiterspace} {(q - 1)}}} } } \right)^{- 1}$.
An adaptive method to determine the value of $\alpha$ is to set it to a fraction $0 \leq \delta < 0.5$ of the distance between the nearest two centers, i.e.\ $\alpha = \delta \min\limits_{i \ne j} {\left\| {\bv_i - \bv_j } \right\|_2^2}$. Following \cite{Ozdemir02}, the fraction value was set to $\delta = 0.4$.

 \item \textbf{Finite-state k-means (FKM)} \cite{Huang04}: This method is a fast approximation for KM. The first iteration is the same as that of KM. In each of the subsequent iterations, the nearest center for a point $\bx_i$ is determined from among the $K'$ ($K' \ll K$) nearest neighbors of the center that the point was assigned to in the previous iteration. When compared to KM, this technique leads to considerable computational savings since the nearest center search is performed in a significantly smaller set of $K'$ centers rather than the entire set of $K$ centers. Following \cite{Huang04}, the number of nearest neighbors was set to $K' = 8$.

 \item \textbf{Stable-flags k-means (SKM)} \cite{Hu07}: This method is another fast approximation for KM. The first $I'$ iterations are the same as those of KM. In the subsequent iterations, the clustering procedure is accelerated using the concepts of center stability and point activity. More specifically, if a cluster center $\bc_k$ does not move by more than $\theta$ units (as measured by the $L_2^2$ distance) in two successive iterations, this center is classified as stable. Furthermore, points that were previously assigned to the stable centers are classified as inactive. At each iteration, only unstable centers and active points participate in the clustering procedure. Following \cite{Hu07}, the algorithm parameters were set to $I' = 10$ and $\theta = 1.0$.
\end{itemize}

For each KM-based quantization method (except for SKM), two variants were implemented. In the first one, the number of iterations was limited to 10, which makes this variant suitable for time-critical applications. These \emph{fixed-iteration} variants are denoted by the plain acronyms KM, FKM, and WSM. In the second variant, to obtain higher quality results, the method was executed until it converged. Convergence was determined by the following commonly used criterion \cite{Linde80}: $ {{\left( {\mbox{SSE}_{i - 1} - \mbox{SSE}_i} \right)} \mathord{\left/ {\vphantom {{\left( {\mbox{SSE}_{i - 1} - \mbox{SSE}_i} \right)} {\mbox{SSE}_i}}} \right. \kern-\nulldelimiterspace} {\mbox{SSE}_i}} \leq \varepsilon $, where $\mbox{SSE}_i$ denotes the SSE \eqref{eq_sse} value at the end of the $i$-th iteration. Following \cite{Huang04,Hu07}, the convergence threshold was set to $\varepsilon = 0.0001$. The \emph{convergent} variants of KM, FKM, and WSM are denoted by KM-C, FKM-C, and WSM-C, respectively. Note that since SKM involves at least $I'=10$ iterations, only the convergent variant was implemented for this method. As for the fuzzy quantization methods, i.e.\ FCM and PIM, due to their excessive computational requirements, the number of iterations for these methods was limited to 10.
\par
Tables \ref{tab_mse}-\ref{tab_time} compare the performance of the methods at quantization levels $K = \{ 32, 64, 128, 256 \}$ on the test images. Note that, for computational simplicity, random initialization was used in the implementations of FCM, PIM, KM, KM-C, FKM, FKM-C, SKM, WSM, and WSM-C. Therefore, in Table \ref{tab_mse}, the quantization errors for these methods are specified in the form of mean ($\mu$) and standard deviation ($\sigma$) over 100 runs. The best (lowest) error values are shown in \textbf{bold}. In addition, with respect to each performance criterion, the methods are ranked based on their mean values over the test images. Table \ref{tab_perf_rank} gives the mean ranks of the methods. The last column gives the overall mean ranks with the assumption that each criterion has equal importance. Note that the best possible rank is 1. The following observations are in order:

\begin{itemize}
\renewcommand{\labelitemi}{$\triangleright$}
 \item In general, the postclustering methods are more effective but less efficient when compared to the preclustering methods.
 \item With respect to distortion minimization, WSM-C outperforms the other methods by a large margin. This method obtains an MSE rank of 1.06, which means that it almost always obtains the lowest distortion.
 \item WSM obtains a significantly better MSE rank than its fixed-iteration rivals.
 \item Overall, WSM and WSM-C are the best methods.
 \item In general, the fastest method is MC, which is followed by SAM, WAN, and WU. The slowest methods are KM-C, FCM, PIM, FKM-C, KM, and SKM.
 \item WSM-C is significantly faster than its convergent rivals. In particular, it provides up to 392 times speed up over KM-C with an average of 62.
 \item WSM is the fastest post-clustering method. It provides up to 46 times speed up over KM with an average of 14.
 \item KM-C, FKM-C, and WSM-C are significantly more stable (particularly when $K$ is small) than their fixed-iteration counterparts as evidenced by their low standard deviation values in Table \ref{tab_mse}. This was expected since these methods were allowed to run longer which helped them overcome potentially adverse initial conditions.
\end{itemize}

Table \ref{tab_stable_rank} gives the mean stability ranks of the methods that involve random initialization. Given a test image and $K$ value combination, the stability of a method is calculated based on the coefficient of variation ($\sigma / \mu$) as: $100 ( 1 - \sigma / \mu )$, where $\mu$ and $\sigma$ denote the mean and standard deviation over 100 runs, respectively. Note that the $\mu$ and $\sigma$ values are given in Table \ref{tab_mse}. Clearly, the higher the stability of a method the better. For example, when $K=32$, WSM-C obtains a mean MSE of 57.461492 with a standard deviation of 0.861126 on the Airplane image. Therefore, the stability of WSM-C in this case is calculated as $100 (1 - 0.861126 / 57.461492) = 98.50$\%. It can be seen that WSM-C is the most stable method, whereas WSM is the most stable fixed-iteration method.
\par
\begin{figure}[!ht]
\centering
 \subfigure[MMM output]{\label{airplane_a}\includegraphics[width=0.2\columnwidth]{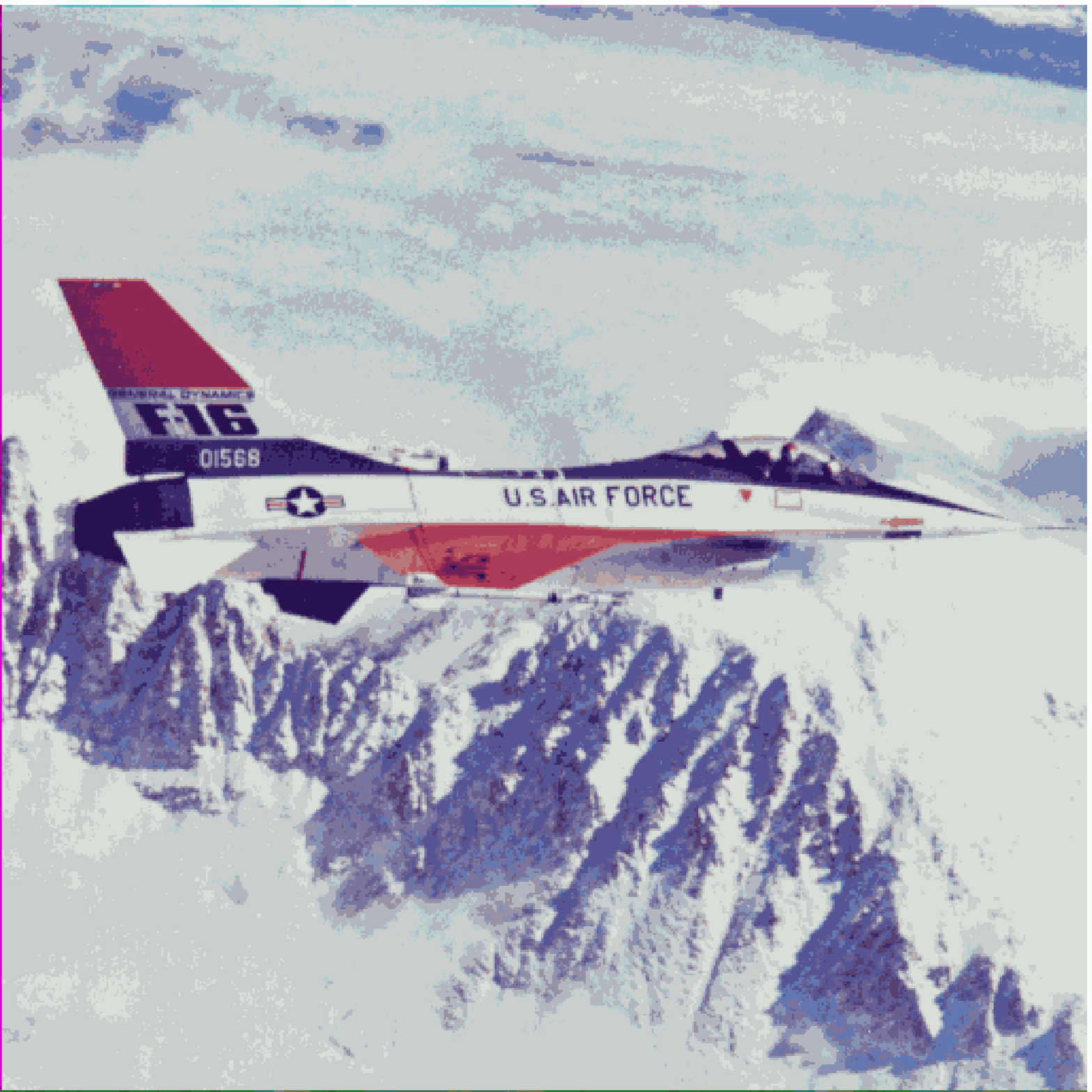}}
 \subfigure[MMM error]{\label{airplane_b}\includegraphics[width=0.2\columnwidth]{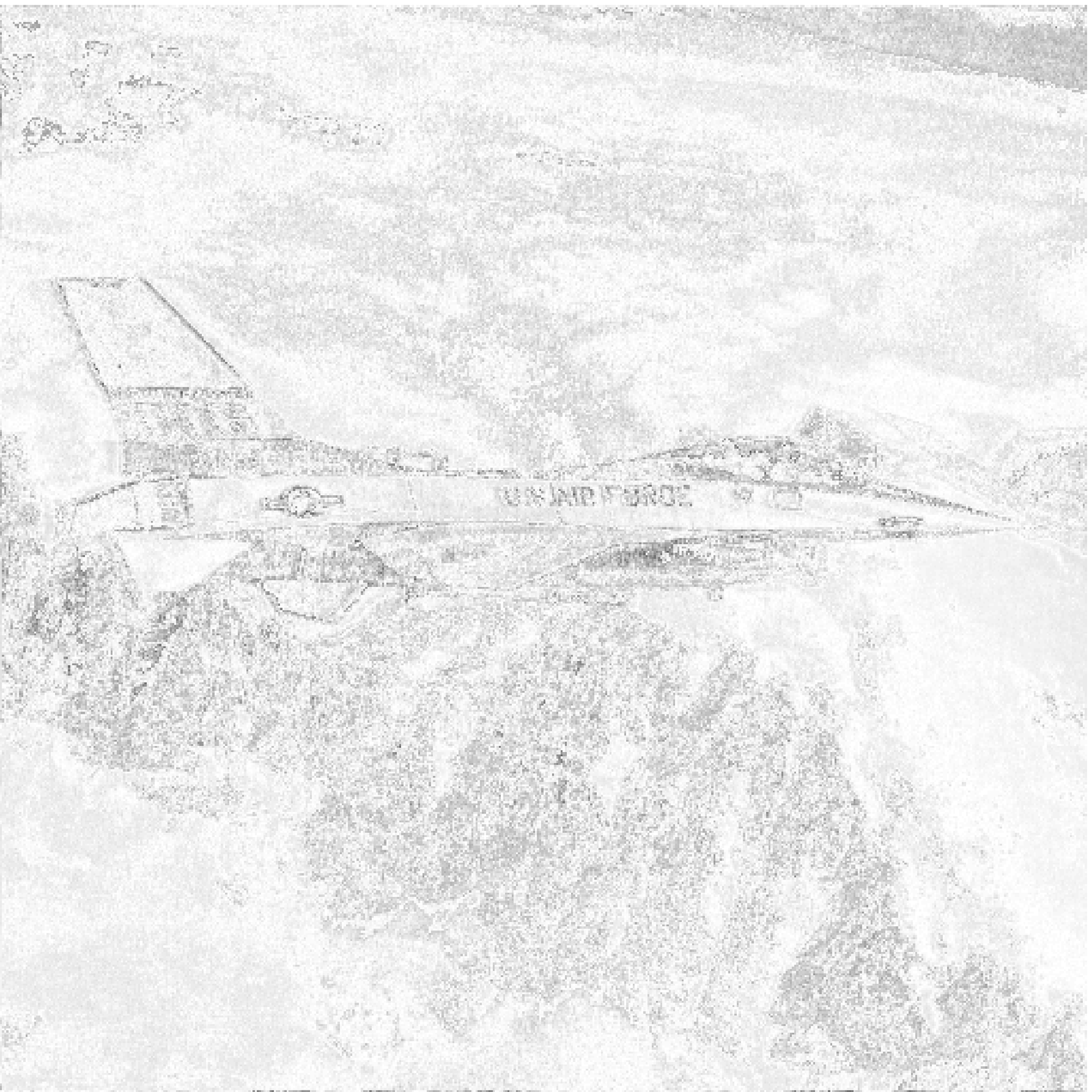}}
 \subfigure[NEU output]{\label{airplane_c}\includegraphics[width=0.2\columnwidth]{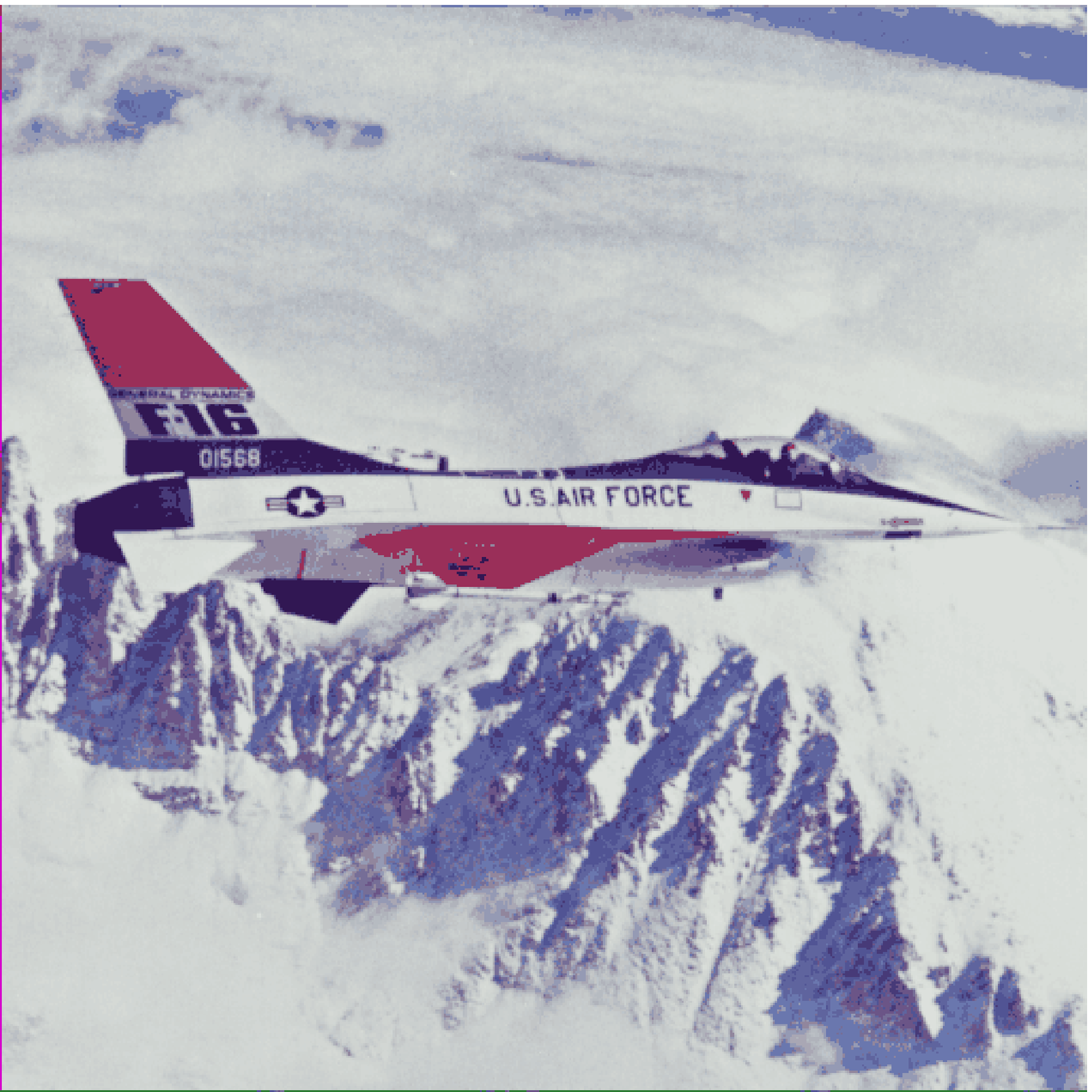}}
 \subfigure[NEU error]{\label{airplane_d}\includegraphics[width=0.2\columnwidth]{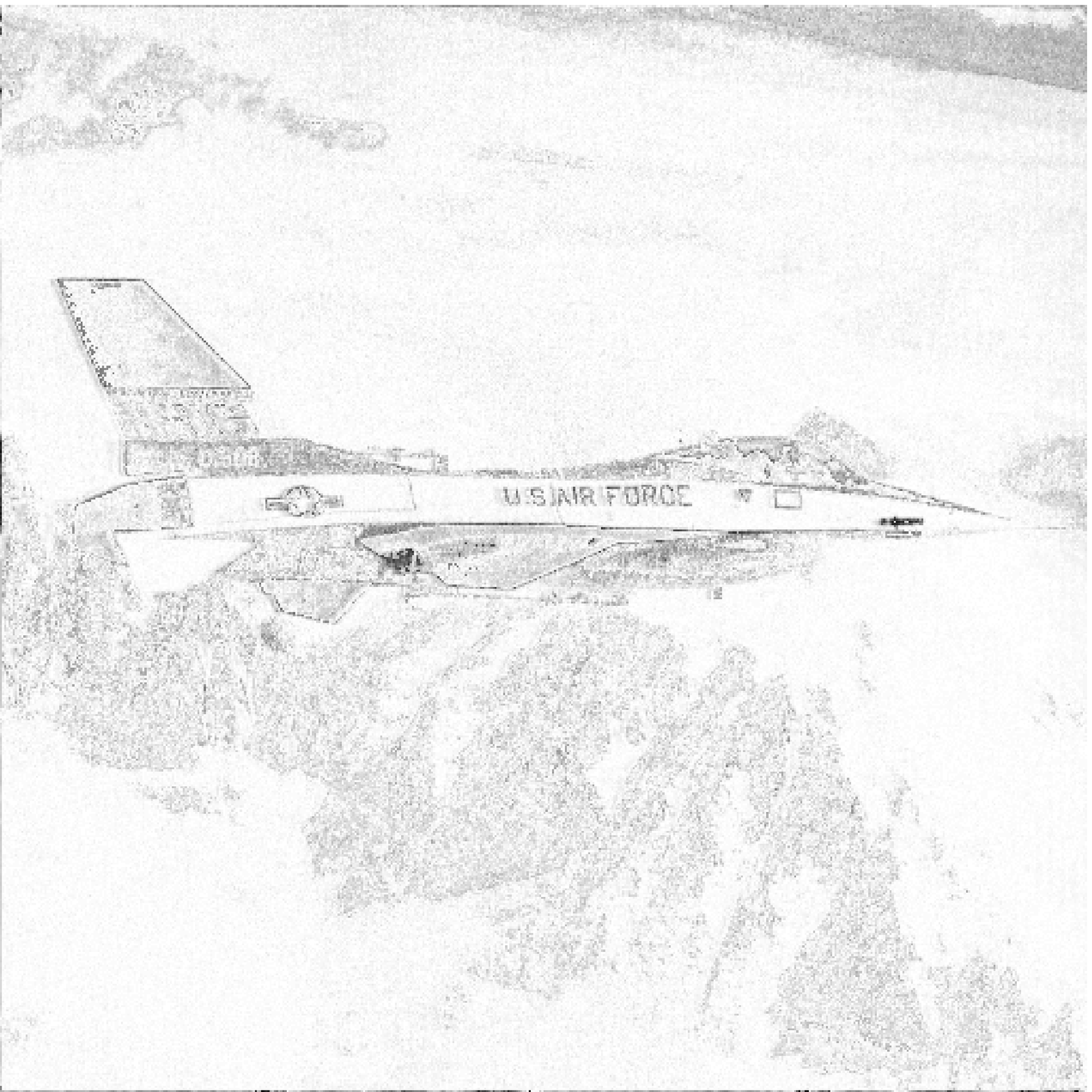}}
 \subfigure[WSM output]{\label{airplane_e}\includegraphics[width=0.2\columnwidth]{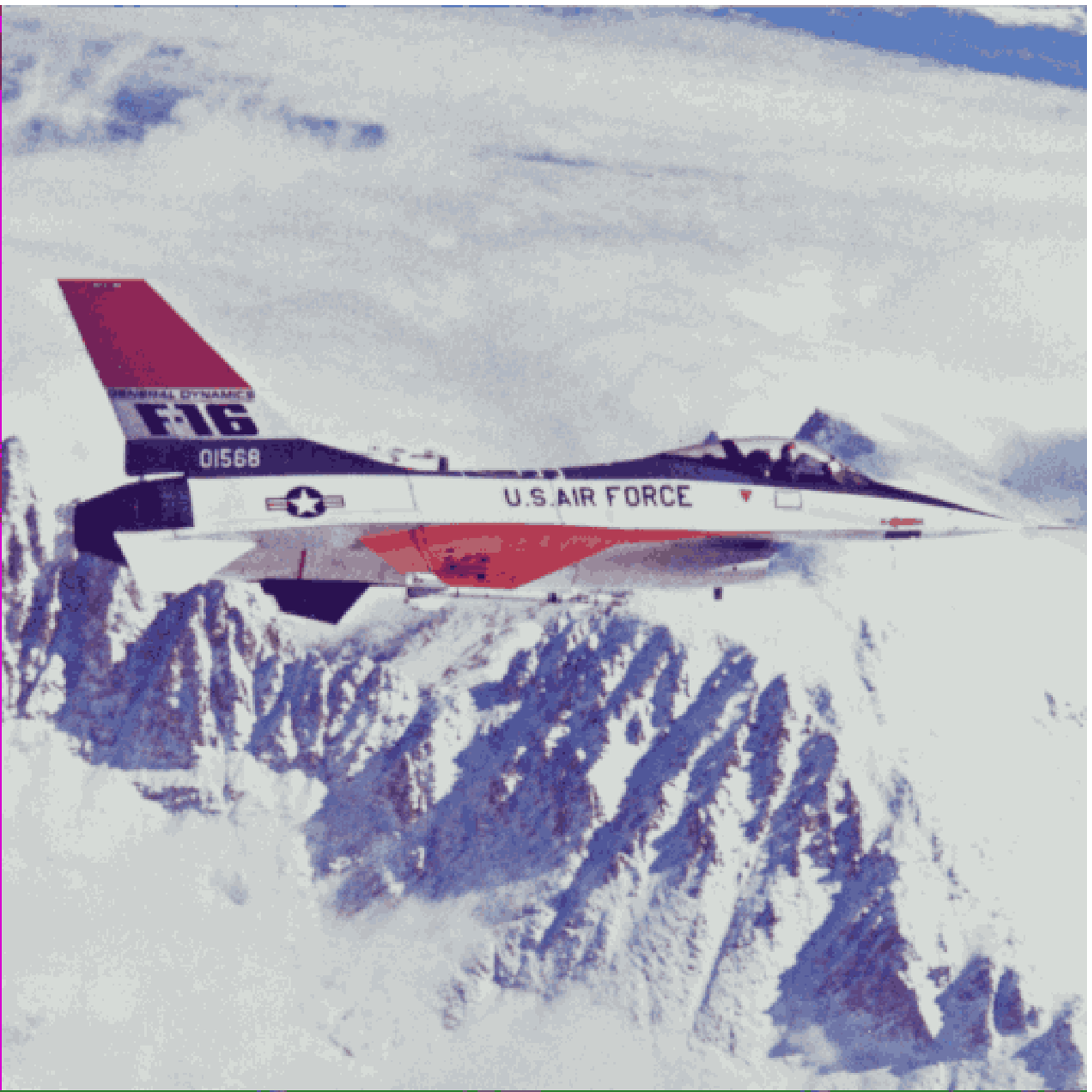}}
 \subfigure[WSM error]{\label{airplane_f}\includegraphics[width=0.2\columnwidth]{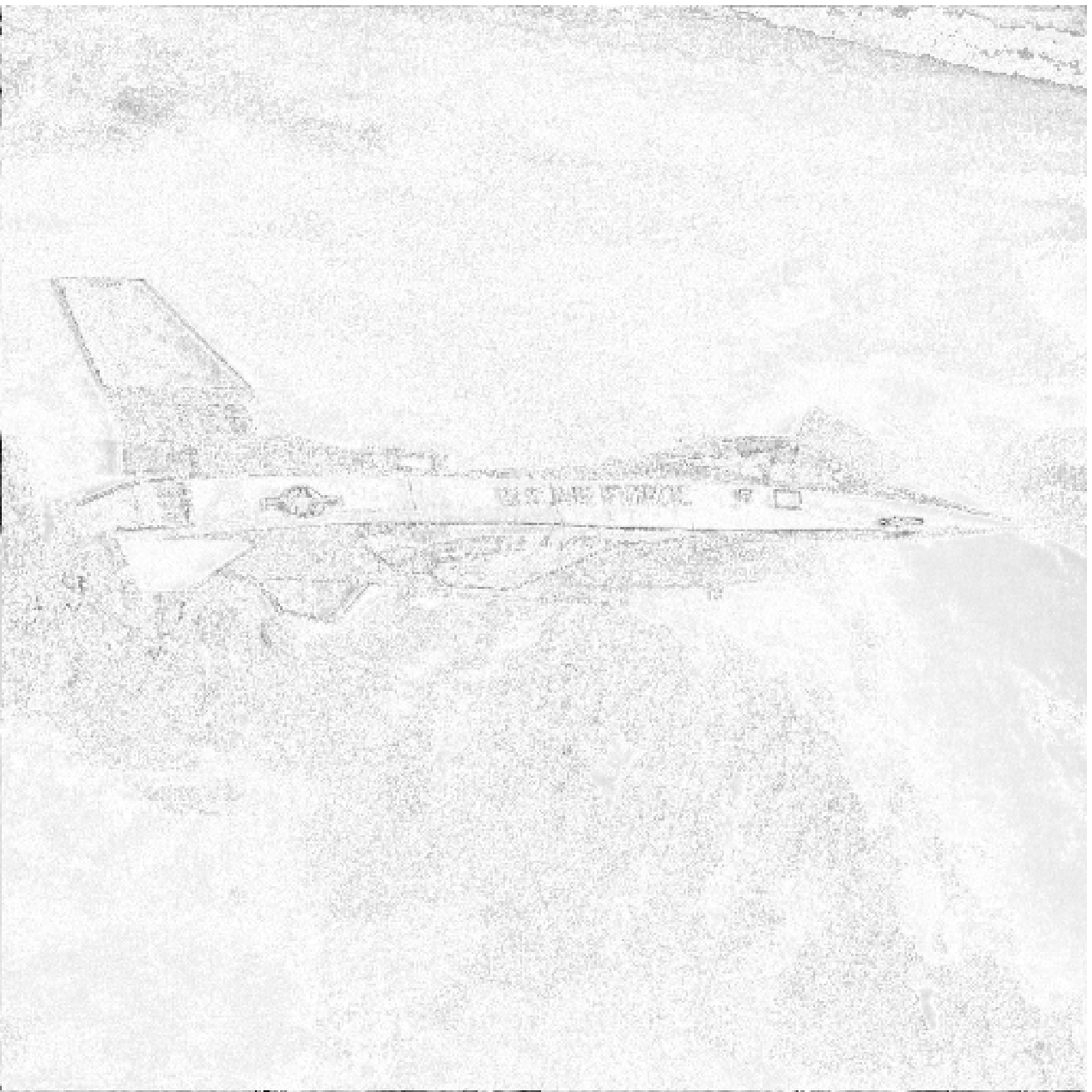}}
 \subfigure[WSM-C output]{\label{airplane_g}\includegraphics[width=0.2\columnwidth]{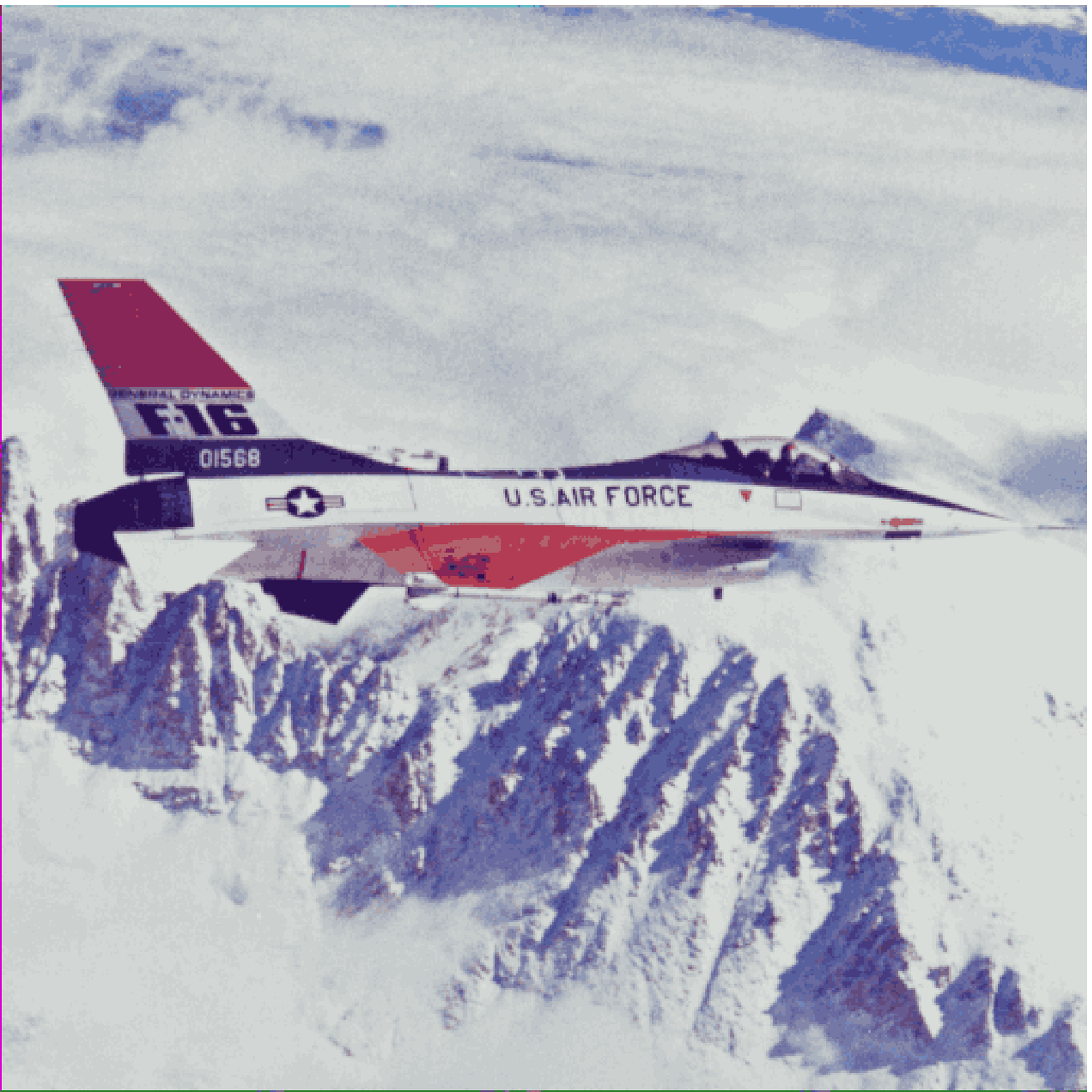}}
 \subfigure[WSM-C error]{\label{airplane_h}\includegraphics[width=0.2\columnwidth]{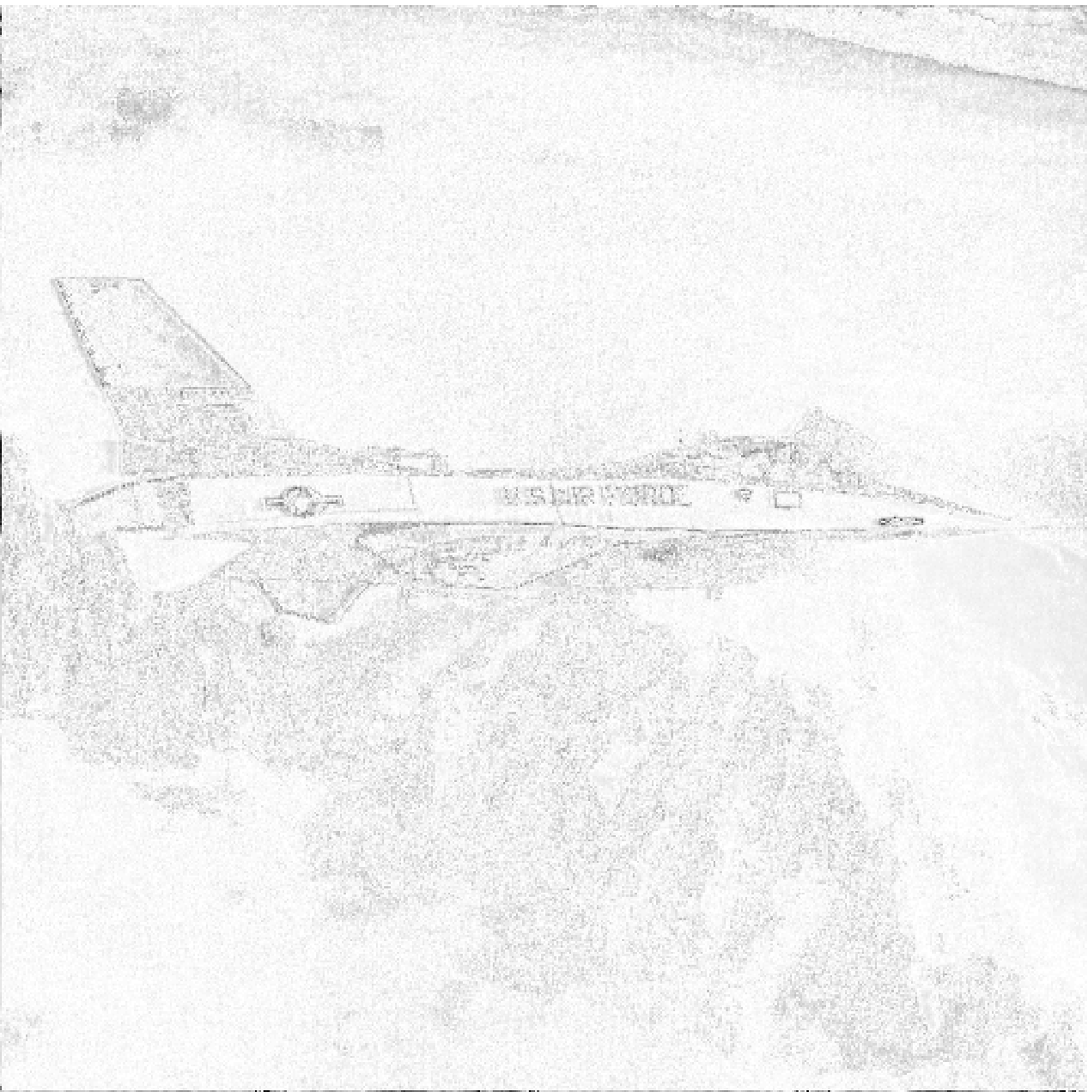}}
 \caption{Sample quantization results for the Airplane image (K=32)}
 \label{fig_airplane}
\end{figure}

Figure \ref{fig_airplane} shows sample quantization results and the corresponding error images. The error image for a particular quantization method was obtained by taking the pixelwise absolute difference between the original and quantized images. In order to obtain a better visualization, pixel values of the error images were multiplied by 4 and then negated. It can be seen that WSM-C and WSM obtain visually pleasing results with less prominent contouring. Furthermore, they achieve the highest color fidelity which is evident by the clean error images that they produce.

\begin{figure}[!ht]
\centering
\includegraphics[width=0.6\columnwidth,draft=false]{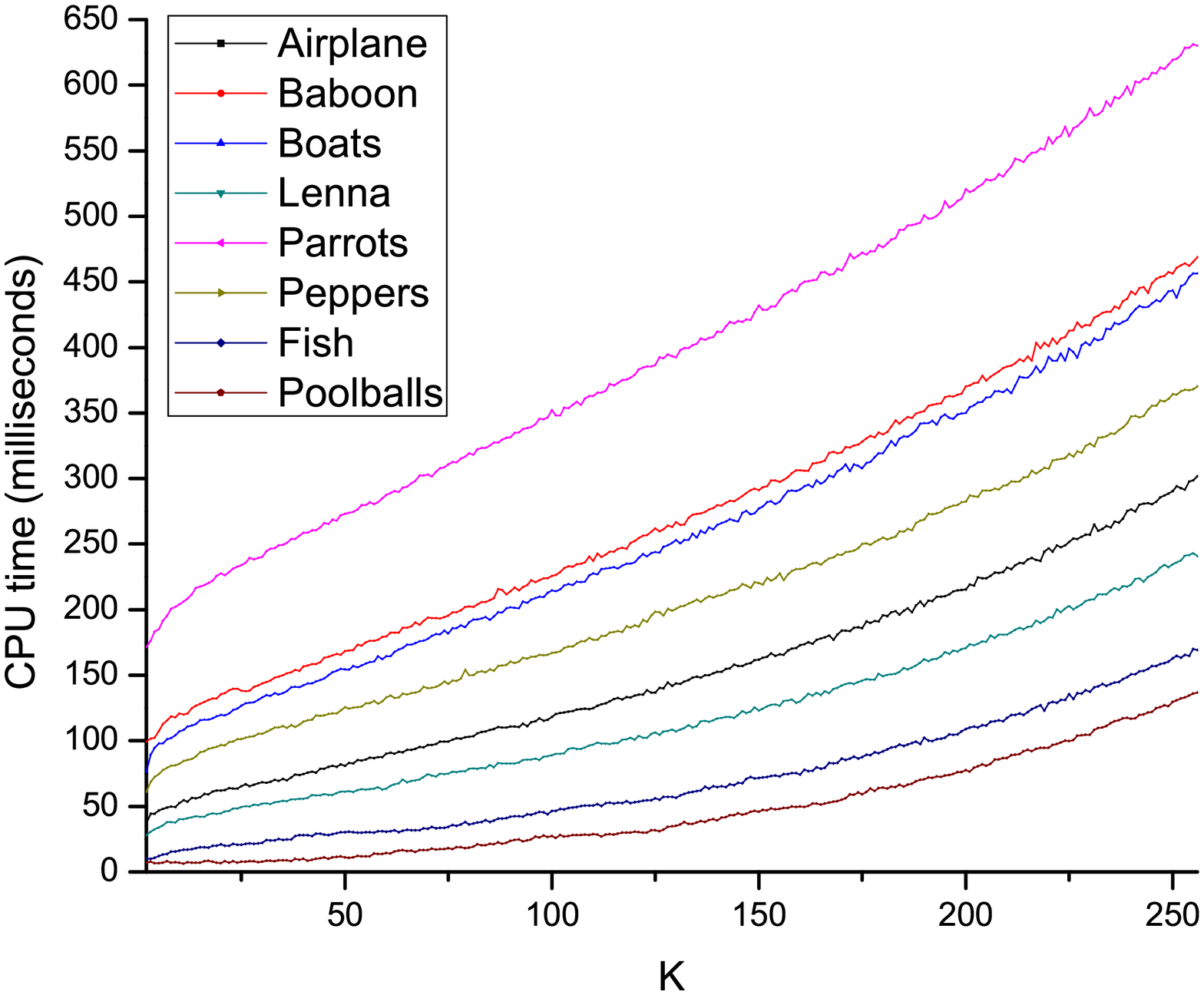}
\caption { \label{fig_graph} CPU time for WSM for $K=\left\{ 2, \ldots, 256 \right\}$ }
\end{figure}

Figure \ref{fig_graph} illustrates the scaling behavior of WSM with respect to $K$. It can be seen that the complexity of WSM is sublinear in $K$, which is due to the intelligent use of the triangle inequality that avoids many distance computations once the cluster centers stabilize after a few iterations. For example, on the Parrots image, increasing $K$ from 2 to 256, results in only about 3.67 fold increase in the computational time (172 ms.\ vs.\ 630 ms.).

We should also mention two other KM-based quantization methods \cite{Kasuga00,Hu08}. As in the case of FKM and SKM, these methods aim to accelerate KM without degrading its effectiveness. However, they do not address the stability problems of KM and thus provide almost the same results in terms of quality. In contrast, WSM (WSM-C) not only provides considerable speed up over KM (KM-C), but also gives significantly better results especially at lower quantization levels.

\section{Conclusions}
\label{sec_conc}
In this paper, a fast and effective color quantization method called WSM (Weighted Sort-Means) was introduced. The method involves several modifications to the conventional k-means algorithm including data reduction, sample weighting, and the use of triangle inequality to speed up the nearest neighbor search. Two variants of WSM were implemented. Although both have very reasonable computational requirements, the fixed-iteration variant is more appropriate for time-critical applications, while the convergent variant should be preferred in applications where obtaining the highest output quality is of prime importance, or the number of quantization levels or the number of unique colors in the original image is small. Experiments on a diverse set of images demonstrated that the two variants of WSM outperform state-of-the-art quantization methods with respect to distortion minimization. Future work will be directed toward the development of a more effective initialization method for WSM.
\par
The implementation of WSM will be made publicly available as part of the Fourier image processing and analysis library, which can be downloaded from \url{http://sourceforge.net/projects/fourier-ipal}.

\section*{Acknowledgments}
This publication was made possible by a grant from The Louisiana Board of Regents (LEQSF2008-11-RD-A-12). The author is grateful to Luiz Velho for the Fish image and Anthony Dekker for the Poolballs image.


\begin{thebibliography}{}
\newcommand{\enquote}[1]{``#1''}

\bibitem{Brun02}
L.~Brun and A.~Tr\'{e}meau, \emph{{Digital Color Imaging Handbook}} (CRC Press,
  2002), chap. {Color Quantization}, pp. 589--638.

\bibitem{Yang98}
C.~-K.~Yang and W.~-H.~Tsai, \enquote{{Color Image Compression Using
  Quantization, Thresholding, and Edge Detection Techniques All Based on the
  Moment-Preserving Principle},} Pattern Recognition Letters \textbf{19},
  205--215 (1998).

\bibitem{Deng01a}
Y.~Deng and B.~Manjunath, \enquote{{Unsupervised Segmentation of Color-Texture
  Regions in Images and Video},} IEEE Trans.\ on Pattern Analysis and Machine
  Intelligence \textbf{23}, 800--810 (2001).

\bibitem{Sertel08}
O.~Sertel, J.~Kong, G.~Lozanski, A.~Shanaah, U.~Catalyurek, J.~Saltz, and
  M.~Gurcan, \enquote{{Texture Classification Using Nonlinear Color
  Quantization: Application to Histopathological Image Analysis},} in
  \enquote{Proc.\ of the IEEE Int.\ Conf.\ on Acoustics, Speech and Signal
  Processing,}  (2008), pp. 597--600.

\bibitem{Kuo07}
C.~-T.~Kuo and S.~-C.~Cheng, \enquote{{Fusion of Color Edge Detection and Color
  Quantization for Color Image Watermarking Using Principal Axes Analysis},}
  Pattern Recognition \textbf{40}, 3691--3704 (2007).

\bibitem{Deng01b}
Y.~Deng, B.~Manjunath, C.~Kenney, M.~Moore, and H.~Shin, \enquote{{An Efficient
  Color Representation for Image Retrieval},} IEEE Trans.\ on Image Processing
  \textbf{10}, 140--147 (2001).

\bibitem{Xiang07}
Z.~Xiang, \emph{{Handbook of Approximation Algorithms and Metaheuristics}}
  (Chapman \& Hall/CRC, 2007), chap. {Color Quantization}, pp. 86--1--86--17.

\bibitem{Gentile90}
R.~S.~Gentile, J.~P.~Allebach, and E.~Walowit, \enquote{{Quantization of Color
  Images Based on Uniform Color Spaces},} Journal of Imaging Technology
  \textbf{16}, 11--21 (1990).

\bibitem{Heckbert82}
P.~Heckbert, \enquote{{Color Image Quantization for Frame Buffer Display},} ACM
  SIGGRAPH Computer Graphics \textbf{16}, 297--307 (1982).

\bibitem{Gervautz88}
M.~Gervautz and W.~Purgathofer, \emph{{New Trends in Computer Graphics}}
  (Springer-Verlag, 1988), chap. {A Simple Method for Color Quantization:
  Octree Quantization}, pp. 219--231.

\bibitem{Wan90}
S.~Wan, P.~Prusinkiewicz, and S.~Wong, \enquote{{Variance-Based Color Image
  Quantization for Frame Buffer Display},} Color Research and Application
  \textbf{15}, 52--58 (1990).

\bibitem{Orchard91}
M.~Orchard and C.~Bouman, \enquote{{Color Quantization of Images},} IEEE
  Trans.\ on Signal Processing \textbf{39}, 2677--2690 (1991).

\bibitem{Wu91}
X.~Wu, \emph{{Graphics Gems Volume II}} (Academic Press, 1991), chap.
  {Efficient Statistical Computations for Optimal Color Quantization}, pp.
  126--133.

\bibitem{Joy93}
G.~Joy and Z.~Xiang, \enquote{{Center-Cut for Color Image Quantization},} The
  Visual Computer \textbf{10}, 62--66 (1993).

\bibitem{Yang96}
C.~-Y.~Yang and J.~-C.~Lin, \enquote{{RWM-Cut for Color Image Quantization},}
  Computers and Graphics \textbf{20}, 577--588 (1996).

\bibitem{Cheng01}
S.~Cheng and C.~Yang, \enquote{{Fast and Novel Technique for Color Quantization
  Using Reduction of Color Space Dimensionality},} Pattern Recognition Letters
  \textbf{22}, 845--856 (2001).

\bibitem{Sirisathitkul04}
Y.~Sirisathitkul, S.~Auwatanamongkol, and B.~Uyyanonvara, \enquote{{Color Image
  Quantization Using Distances between Adjacent Colors along the Color Axis
  with Highest Color Variance},} Pattern Recognition Letters \textbf{25},
  1025--1043 (2004).

\bibitem{Kanjanawanishkul05}
K.~Kanjanawanishkul and B.~Uyyanonvara, \enquote{{Novel Fast Color Reduction
  Algorithm for Time-Constrained Applications},} Journal of Visual
  Communication and Image Representation \textbf{16}, 311--332 (2005).

\bibitem{Equitz89}
W.~H.~Equitz, \enquote{{A New Vector Quantization Clustering Algorithm},} IEEE
  Trans.\ on Acoustics, Speech and Signal Processing \textbf{37}, 1568--1575
  (1989).

\bibitem{Balasubramanian91}
R.~Balasubramanian and J.~Allebach, \enquote{{A New Approach to Palette
  Selection for Color Images},} Journal of Imaging Technology \textbf{17},
  284--290 (1991).

\bibitem{Xiang94}
Z.~Xiang and G.~Joy, \enquote{{Color Image Quantization by Agglomerative
  Clustering},} IEEE Computer Graphics and Applications \textbf{14}, 44--48
  (1994).

\bibitem{Velho97}
L.~Velho, J.~Gomez, and M.~Sobreiro, \enquote{{Color Image Quantization by
  Pairwise Clustering},} in \enquote{Proc.\ of the 10th Brazilian Symposium on
  Computer Graphics and Image Processing,}  (1997), pp. 203--210.

\bibitem{Brun00}
L.~Brun and M.~Mokhtari, \enquote{{Two High Speed Color Quantization
  Algorithms},} in \enquote{Proc.\ of the 1st Int.\ Conf.\ on Color in Graphics
  and Image Processing,}  (2000), pp. 116--121.

\bibitem{Kasuga00}
H.~Kasuga, H.~Yamamoto, and M.~Okamoto, \enquote{{Color Quantization Using the
  Fast K-Means Algorithm},} Systems and Computers in Japan \textbf{31}, 33--40
  (2000).

\bibitem{Huang04}
Y.-L. Huang and R.-F. Chang, \enquote{{A Fast Finite-State Algorithm for
  Generating RGB Palettes of Color Quantized Images},} Journal of Information
  Science and Engineering \textbf{20}, 771--782 (2004).

\bibitem{Hu07}
Y.-C. Hu and M.-G. Lee, \enquote{{K-means Based Color Palette Design Scheme
  with the Use of Stable Flags},} Journal of Electronic Imaging \textbf{16},
  033003 (2007).

\bibitem{Hu08}
Y.-C. Hu and B.-H. Su, \enquote{{Accelerated K-means Clustering Algorithm for
  Colour Image Quantization},} Imaging Science Journal \textbf{56}, 29--40
  (2008).

\bibitem{Xiang97}
Z.~Xiang, \enquote{{Color Image Quantization by Minimizing the Maximum
  Intercluster Distance},} ACM Trans.\ on Graphics \textbf{16}, 260--276
  (1997).

\bibitem{Verevka95}
O.~Verevka and J.~Buchanan, \enquote{{Local K-Means Algorithm for Colour Image
  Quantization},} in \enquote{Proc.\ of the Graphics/Vision Interface Conf.},
  (1995), pp. 128--135.

\bibitem{Scheunders97}
P.~Scheunders, \enquote{{Comparison of Clustering Algorithms Applied to Color
  Image Quantization},} Pattern Recognition Letters \textbf{18}, 1379--1384
  (1997).

\bibitem{Celebi09}
M.E. Celebi, \enquote{{An Effective Color Quantization Method Based on the Competitive
	Learning Paradigm},} in \enquote{Proc.\ of the Int.\ Conf.\ on Image Processing, Computer Vision,
	and Pattern Recognition}, (2009), pp. 876--880.

\bibitem{Ozdemir02}
D.~Ozdemir and L.~Akarun, \enquote{{Fuzzy Algorithm for Color Quantization of
  Images},} Pattern Recognition \textbf{35}, 1785--1791 (2002).

\bibitem{Schaefer09}
G.~Schaefer and H.~Zhou, \enquote{{Fuzzy Clustering for Colour Reduction in
  Images},} Telecommunication Systems \textbf{40}, 17--25 (2009).

\bibitem{Bing04}
Z.~Bing, S.~Junyi, and P.~Qinke, \enquote{{An Adjustable Algorithm for Color
  Quantization},} Pattern Recognition Letters \textbf{25}, 1787--–1797 (2004).

\bibitem{Dekker94}
A.~Dekker, \enquote{{Kohonen Neural Networks for Optimal Colour Quantization},}
  Network: Computation in Neural Systems \textbf{5}, 351--367 (1994).

\bibitem{Papamarkos02}
N.~Papamarkos, A.~Atsalakis, and C.~Strouthopoulos, \enquote{{Adaptive Color
  Reduction},} IEEE Trans.\ on Systems, Man, and Cybernetics Part B
  \textbf{32}, 44--56 (2002).

\bibitem{Chang05}
C.-H.~Chang, P.~Xu, R.~Xiao, and T.~Srikanthan, \enquote{{New Adaptive Color
  Quantization Method Based on Self-Organizing Maps},} IEEE Trans.\ on Neural
  Networks \textbf{16}, 237--249 (2005).

\bibitem{Linde80}
Y.~Linde, A.~Buzo, and R.~Gray, \enquote{{An Algorithm for Vector Quantizer
  Design},} IEEE Trans.\ on Communications \textbf{28}, 84--95 (1980).

\bibitem{Gan07}
G.~Gan, C.~Ma, and J.~Wu, \emph{{Data Clustering: Theory, Algorithms, and
  Applications}} (SIAM, 2007).

\bibitem{Drineas04}
P.~Drineas, A.~Frieze, R.~Kannan, S.~Vempala, and V.~Vinay,
  \enquote{{Clustering Large Graphs via the Singular Value Decomposition},}
  Machine Learning \textbf{56}, 9–--33 (2004).

\bibitem{Lloyd82}
S.~Lloyd, \enquote{{Least Squares Quantization in PCM},} IEEE Trans.\ on
  Information Theory \textbf{28}, 129--136 (1982).

\bibitem{Forgy65}
E.~Forgy, \enquote{{Cluster Analysis of Multivariate Data: Efficiency vs.\
  Interpretability of Classification},} Biometrics \textbf{21}, 768 (1965).

\bibitem{Phillips02}
S.~Phillips, \enquote{{Acceleration of K-Means and Related Clustering
  Algorithms},} in \enquote{Proc.\ of the 4th Int.\ Workshop on Algorithm
  Engineering and Experiments,}  (2002), pp. 166--177.

\bibitem{Kanungo02}
T.~Kanungo, D.~Mount, N.~Netanyahu, C.~Piatko, R.~Silverman, and A.~Wu,
  \enquote{{An Efficient K-Means Clustering Algorithm: Analysis and
  Implementation},} IEEE Trans.\ on Pattern Analysis and Machine Intelligence
  \textbf{24}, 881--892 (2002).

\bibitem{Har-Peled04}
S.~Har-Peled and A.~Kushal, \enquote{{Smaller Coresets for K-Median and K-Means
  Clustering},} in \enquote{Proc.\ of the 21st Annual Symposium on
  Computational Geometry,}  (2004), pp. 126--134.

\bibitem{Elkan03}
C.~Elkan, \enquote{{Using the Triangle Inequality to Accelerate K-Means},} in
  \enquote{Proc.\ of the 20th Int.\ Conf.\ on Machine Learning,}  (2003), pp.
  147--153.

\bibitem{Bezdek81}
J.~C.~Bezdek, \emph{{Pattern Recognition with Fuzzy Objective Function
  Algorithms}} (Springer-Verlag, 1981).

\bibitem{Kolen02}
J.~F.~Kolen and T.~Hutcheson, \enquote{{Reducing the Time Complexity of the
  Fuzzy C-Means Algorithm},} IEEE Trans.\ on Fuzzy Systems \textbf{10},
  263--267 (2002).

\end{thebibliography}

\newpage

\begin{table}[ht]
\linespread{1}
\centering
\scriptsize
{
\caption{ \label{tab_mse} MSE comparison of the quantization methods}
\begin{tabular}{|c|cc|cc|cc|cc||cc|cc|cc|cc|}
\hline
 & \multicolumn{2}{c|}{K = 32} & \multicolumn{2}{c|}{K = 64} & \multicolumn{2}{c|}{K = 128} & \multicolumn{2}{c||}{K = 256} & \multicolumn{2}{c|}{K = 32} & \multicolumn{2}{c|}{K = 64} & \multicolumn{2}{c|}{K = 128} & \multicolumn{2}{c|}{K = 256}\\
 Method & $\mu$ & $\sigma$ & $\mu$ & $\sigma$ & $\mu$ & $\sigma$ & $\mu$ & $\sigma$ &
 $\mu$ & $\sigma$ & $\mu$ & $\sigma$ & $\mu$ & $\sigma$ & $\mu$ & $\sigma$\\
\hline
\multicolumn{1}{|c|}{} & \multicolumn{8}{|c||}{Airplane} & \multicolumn{8}{|c|}{Baboon}\\
\hline
MC & 124 & - & 81 & - & 54 & - & 41 & - & 546 & - & 371 & - & 248 & - & 166 & -\\
WAN & 117 & - & 69 & - & 50 & - & 39 & - & 509 & - & 326 & - & 216 & - & 142 & -\\
WU & 75 & - & 47 & - & 30 & - & 21 & - & 422 & - & 248 & - & 155 & - & 99 & -\\
NEU & 101 & - & 47 & - & 24 & - & 15 & - & 363 & - & 216 & - & 128 & - & 84 & -\\
MMM & 134 & - & 82 & - & 44 & - & 28 & - & 489 & - & 270 & - & 189 & - & 120 & -\\
SAM & 120 & - & 65 & - & 43 & - & 31 & - & 396 & - & 245 & - & 153 & - & 99 & - \\
FCM & 74 & 9 & 44 & 4 & 29 & 2 & 21 & 1 & 415 & 15 & 265 & 10 & 174 & 6 & 119 & 4\\
PIM & 73 & 9 & 45 & 4 & 29 & 2 & 21 & 1 & 413 & 18 & 261 & 13 & 172 & 7 & 117 & 4\\
KM & 112 & 25 & 65 & 12 & 36 & 4 & 22 & 2 & 345 & 9 & 206 & 5 & 129 & 2 & 83 & 1\\
KM-C & 59 & 2 & 35 & 1 & 25 & 0 & 19 & 0 & 329 & 3 & 196 & 1 & \textbf{123} & 1 & 79 & 0\\
FKM & 113 & 19 & 64 & 9 & 36 & 4 & 22 & 1 & 346 & 9 & 206 & 4 & 129 & 2 & 83 & 1\\
FKM-C & 59 & 2 & 35 & 1 & 26 & 1 & 19 & 1 & 328 & 3 & 196 & 1 & \textbf{123} & 1 & 79 & 0\\
SKM & 112 & 20 & 63 & 9 & 36 & 4 & 22 & 1 & 343 & 10 & 207 & 6 & 129 & 2 & 83 & 1\\
\emph{WSM} & 64 & 4 & 36 & 1 & 23 & 1 & 15 & 0 & 345 & 8 & 204 & 3 & 127 & 1 & 81 & 1\\
\emph{WSM-C} & \textbf{57} & 1 & \textbf{34} & 0 & \textbf{22} & 0 & \textbf{14} & 0 & \textbf{327} & 3 & \textbf{195} & 1 & \textbf{123} & 1 & \textbf{78} & 0\\
\hline
\multicolumn{1}{|c|}{} & \multicolumn{8}{|c||}{Boats} & \multicolumn{8}{|c|}{Lenna}\\
\hline
MC & 200 & - & 126 & - & 78 & - & 57 & - & 165 & - & 94 & - & 71 & - & 47 & -\\
WAN & 198 & - & 117 & - & 71 & - & 45 & - & 159 & - & 93 & - & 61 & - & 43 & -\\
WU & 154 & - & 87 & - & 50 & - & 32 & - & 130 & - & 76 & - & 46 & - & 29 & -\\
NEU & 147 & - & 79 & - & 41 & - & 26 & - & 119 & - & 68 & - & 36 & - & 23 & -\\
MMM & 203 & - & 114 & - & 69 & - & 41 & - & 139 & - & 86 & - & 50 & - & 34 & -\\
SAM & 161 & - & 95 & - & 59 & - & 42 & - & 135 & - & 88 & - & 56 & - & 40 & -\\
FCM & 160 & 13 & 99 & 8 & 64 & 5 & 42 & 3 & 132 & 10 & 83 & 7 & 53 & 4 & 38 & 2\\
PIM & 161 & 14 & 99 & 11 & 63 & 5 & 43 & 3 & 136 & 12 & 81 & 6 & 53 & 4 & 38 & 2\\
KM & 135 & 11 & 78 & 5 & 47 & 3 & 30 & 1 & 106 & 5 & 61 & 2 & 38 & 1 & 24 & 0\\
KM-C & \textbf{115} & 1 & 64 & 1 & 39 & 0 & 25 & 0 & \textbf{97} & 1 & 57 & 1 & 35 & 0 & \textbf{22} & 0\\
FKM & 134 & 10 & 77 & 5 & 47 & 3 & 29 & 1 & 107 & 8 & 61 & 2 & 38 & 1 & 24 & 0\\
FKM-C & 116 & 1 & 65 & 1 & 39 & 0 & 25 & 0 & \textbf{97} & 1 & 57 & 1 & 35 & 0 & \textbf{22} & 0\\
SKM & 137 & 13 & 77 & 4 & 47 & 2 & 30 & 1 & 107 & 6 & 62 & 2 & 38 & 1 & 24 & 1\\
\emph{WSM} & 125 & 7 & 68 & 2 & 40 & 1 & 24 & 0 & 103 & 5 & 60 & 2 & 36 & 1 & 23 & 0\\
\emph{WSM-C} & \textbf{115} & 1 & \textbf{63} & 0 & \textbf{37} & 0 & \textbf{23} & 0 & \textbf{97} & 2 & \textbf{56} & 1 & \textbf{34} & 0 & \textbf{22} & 0\\
\hline
\multicolumn{1}{|c|}{} & \multicolumn{8}{|c||}{Parrots} & \multicolumn{8}{|c|}{Peppers}\\
\hline
MC & 401 & - & 258 & - & 144 & - & 99 & - & 333 & - & 213 & - & 147 & - & 98 & -\\
WAN & 365 & - & 225 & - & 146 & - & 90 & - & 333 & - & 215 & - & 142 & - & 93 & -\\
WU & 291 & - & 171 & - & 96 & - & 59 & - & 264 & - & 160 & - & 101 & - & 63 & -\\
NEU & 306 & - & 153 & - & 84 & - & 47 & - & 249 & - & 151 & - & 83 & - & 55 & -\\
MMM & 332 & - & 200 & - & 117 & - & 73 & - & 292 & - & 182 & - & 113 & - & 76 & -\\
SAM & 276 & - & 160 & - & 94 & - & 60 & - & 268 & - & 161 & - & 100 & - & 64 & - \\
FCM & 297 & 19 & 178 & 14 & 107 & 5 & 69 & 2 & 272 & 15 & 179 & 7 & 120 & 4 & 84 & 3\\
PIM & 295 & 21 & 175 & 12 & 107 & 5 & 69 & 2 & 266 & 14 & 176 & 7 & 119 & 5 & 84 & 3\\
KM & 262 & 20 & 149 & 9 & 85 & 4 & 51 & 2 & 232 & 7 & 141 & 4 & 87 & 2 & 54 & 1\\
KM-C & 237 & 7 & 131 & 3 & 76 & 1 & 46 & 1 & 220 & 2 & 132 & 1 & \textbf{80} & 0 & 51 & 0\\
FKM & 264 & 21 & 150 & 10 & 87 & 4 & 51 & 2 & 231 & 6 & 142 & 4 & 86 & 2 & 55 & 1\\
FKM-C & 237 & 7 & 132 & 3 & 77 & 2 & 47 & 1 & 220 & 2 & 132 & 2 & 81 & 1 & 51 & 0\\
SKM & 259 & 16 & 152 & 11 & 86 & 4 & 51 & 2 & 233 & 7 & 142 & 4 & 87 & 2 & 55 & 1\\
\emph{WSM} & 249 & 13 & 136 & 5 & 79 & 2 & 46 & 1 & 232 & 7 & 139 & 3 & 85 & 1 & 53 & 1\\
\emph{WSM-C} & \textbf{232} & 6 & \textbf{128} & 2 & \textbf{74} & 1 & \textbf{43} & 0 & \textbf{219} & 2 & \textbf{131} & 1 & \textbf{80} & 1 & \textbf{50} & 0\\
\hline
\multicolumn{1}{|c|}{} & \multicolumn{8}{|c||}{Fish} & \multicolumn{8}{|c|}{Poolballs}\\
\hline
MC & 276 & - & 169 & - & 107 & - & 68 & - & 136 & - & 64 & - & 38 & - & 27 & -\\
WAN & 311 & - & 208 & - & 124 & - & 77 & - & 112 & - & 59 & - & 45 & - & 38 & -\\
WU & 187 & - & 111 & - & 69 & - & 44 & - & 68 & - & 31 & - & 17 & - & 11 & -\\
NEU & 173 & - & 107 & - & 57 & - & 42 & - & 104 & - & 44 & - & 18 & - & 9 & -\\
MMM & 235 & - & 136 & - & 81 & - & 53 & - & 166 & - & 91 & - & 42 & - & 20 & -\\
SAM & 198 & - & 120 & - & 74 & - & 49 & - & 91 & - & 54 & - & 37 & - & 20 & -\\
FCM & 169 & 11 & 110 & 5 & 79 & 3 & 60 & 3 & 153 & 75 & 61 & 30 & 25 & 5 & 14 & 2\\
PIM & 168 & 9 & 111 & 4 & 79 & 3 & 60 & 3 & 149 & 71 & 57 & 26 & 25 & 7 & 14 & 2\\
KM & 174 & 24 & 105 & 9 & 64 & 4 & 40 & 2 & 226 & 75 & 129 & 31 & 75 & 17 & 39 & 8\\
KM-C & 145 & 3 & 90 & 2 & 58 & 2 & 37 & 1 & 94 & 8 & 51 & 5 & 44 & 6 & 29 & 5\\
FKM & 173 & 17 & 105 & 10 & 65 & 4 & 40 & 2 & 229 & 73 & 130 & 44 & 78 & 15 & 37 & 6\\
FKM-C & 144 & 3 & 90 & 2 & 59 & 2 & 38 & 1 & 95 & 9 & 55 & 10 & 45 & 8 & 27 & 5\\
SKM & 177 & 19 & 105 & 9 & 65 & 4 & 40 & 2 & 167 & 35 & 120 & 15 & 71 & 13 & 37 & 7\\
\emph{WSM} & 148 & 3 & 91 & 3 & 55 & 1 & 33 & 0 & 69 & 10 & 31 & 6 & 14 & 2 & \textbf{7} & 0\\
\emph{WSM-C} & \textbf{142} & 4 & \textbf{85} & 1 & \textbf{52} & 1 & \textbf{32} & 0 & \textbf{62} & 6 & \textbf{27} & 3 & \textbf{13} & 1 & \textbf{7} & 0\\
\hline
\end{tabular}
}
\end{table}

\begin{table}[ht]
\linespread{1}
\centering
\scriptsize
{
\caption{ \label{tab_time} CPU time comparison of the quantization methods}
\begin{tabular}{|c|c|c|c|c||c|c|c|c|}
\hline
Method & K = 32 & K = 64 & K = 128 & K = 256 & K = 32 & K = 64 & K = 128 & K = 256\\
\hline
\multicolumn{1}{|c|}{} & \multicolumn{4}{|c||}{Airplane} & \multicolumn{4}{|c|}{Baboon}\\
\hline
MC & 10 & 10 & \textbf{11} & \textbf{12} & 10 & \textbf{10} & \textbf{11} & \textbf{13}\\
WAN & 13 & 14 & 15 & 18 & 14 & 15 & 16 & 20\\
WU & 16 & 16 & 16 & 16 & 16 & 15 & 16 & 17\\
NEU & 70 & 142 & 265 & 514 & 67 & 134 & 254 & 485\\
MMM & 123 & 206 & 367 & 696 & 126 & 207 & 375 & 702\\
SAM & \textbf{7} & \textbf{8} & 13 & 25 & \textbf{9} & 20 & 56 & 112\\
FCM & 2739 & 5285 & 10612 & 21079 & 2737 & 5285 & 10612 & 21081\\
PIM & 2410 & 5038 & 10402 & 20913 & 2488 & 5091 & 10407 & 20846\\
KM & 584 & 1005 & 1791 & 3314 & 592 & 1012 & 1800 & 3317\\
KM-C & 17688 & 43850 & 74814 & 71908 & 3136 & 7070 & 13164 & 25657\\
FKM & 189 & 222 & 299 & 505 & 189 & 223 & 299 & 508\\
FKM-C & 4111 & 6144 & 6057 & 5376 & 746 & 934 & 1171 & 1959\\
SKM & 530 & 903 & 1593 & 2952 & 547 & 927 & 1610 & 2961\\
\emph{WSM} & 68 & 92 & 145 & 301 & 147 & 188 & 270 & 477\\
\emph{WSM-C} & 257 & 359 & 522 & 1180 & 401 & 565 & 814 & 1580\\
\hline
\multicolumn{1}{|c|}{} & \multicolumn{4}{|c||}{Boats} & \multicolumn{4}{|c|}{Lenna}\\
\hline
MC & 19 & \textbf{18} & \textbf{19} & \textbf{21} & 9 & 8 & 10 & \textbf{10}\\
WAN & 24 & 24 & 26 & 29 & 12 & 15 & 15 & 17\\
WU & 28 & 26 & 28 & 28 & 15 & 15 & 14 & 15\\
NEU & 122 & 232 & 453 & 853 & 61 & 123 & 244 & 465\\
MMM & 219 & 367 & 656 & 1237 & 116 & 193 & 346 & 654\\
SAM & \textbf{17} & 19 & 21 & 32 & \textbf{8} & \textbf{7} & \textbf{9} & 13\\
FCM & 4695 & 9141 & 18350 & 36471 & 2545 & 4954 & 9953 & 19770\\
PIM & 4075 & 8555 & 17784 & 36071 & 2348 & 4820 & 9832 & 19681\\
KM & 986 & 1727 & 3087 & 5729 & 536 & 939 & 1673 & 3101\\
KM-C & 9853 & 22622 & 53858 & 111047 & 3457 & 6698 & 11927 & 23762\\
FKM & 326 & 385 & 509 & 804 & 170 & 205 & 281 & 478\\
FKM-C & 2393 & 3158 & 4007 & 6056 & 788 & 878 & 1167 & 1886\\
SKM & 908 & 1551 & 2756 & 5105 & 485 & 837 & 1493 & 2778\\
\emph{WSM} & 136 & 174 & 255 & 464 & 52 & 68 & 110 & 244\\
\emph{WSM-C} & 486 & 614 & 853 & 1647 & 149 & 212 & 329 & 883\\
\hline
\multicolumn{1}{|c|}{} & \multicolumn{4}{|c||}{Parrots} & \multicolumn{4}{|c|}{Peppers}\\
\hline
MC & \textbf{57} & \textbf{58} & \textbf{59} & \textbf{61} & 10 & \textbf{10} & \textbf{11} & \textbf{12}\\
WAN & 81 & 82 & 83 & 86 & 13 & 14 & 16 & 18\\
WU & 86 & 87 & 86 & 87 & 16 & 17 & 17 & 17\\
NEU & 476 & 849 & 1571 & 2914 & 70 & 135 & 262 & 493\\
MMM & 758 & 1265 & 2282 & 4286 & 125 & 206 & 371 & 700\\
SAM & 74 & 77 & 103 & 150 & \textbf{8} & 11 & 29 & 53\\
FCM & 16096 & 31734 & 63871 & 126554 & 2739 & 5288 & 10624 & 21107\\
PIM & 14620 & 30159 & 61891 & 124794 & 2499 & 5107 & 10425 & 20883\\
KM & 3309 & 5918 & 10657 & 19828 & 564 & 996 & 1785 & 3309\\
KM-C & 23949 & 61168 & 119907 & 242439 & 3387 & 7761 & 14839 & 31893\\
FKM & 1100 & 1302 & 1698 & 2519 & 181 & 219 & 295 & 500\\
FKM-C & 5464 & 8557 & 9529 & 10482 & 869 & 1017 & 1262 & 2233\\
SKM & 3072 & 5429 & 9506 & 17599 & 523 & 905 & 1605 & 2971\\
\emph{WSM} & 250 & 298 & 399 & 639 & 107 & 138 & 201 & 373\\
\emph{WSM-C} & 634 & 820 & 1261 & 2149 & 327 & 466 & 648 & 1387\\
\hline
\multicolumn{1}{|c|}{} & \multicolumn{4}{|c||}{Fish} & \multicolumn{4}{|c|}{Poolballs}\\
\hline
MC & 6 & \textbf{5} & \textbf{7} & \textbf{6} & \textbf{9} & \textbf{9} & \textbf{9} & \textbf{11}\\
WAN & 5 & 6 & 8 & 12 & 10 & 10 & 12 & 14\\
WU & 8 & 9 & 8 & 9 & 12 & 13 & 12 & 13\\
NEU & 12 & 27 & 58 & 110 & 51 & 103 & 192 & 353\\
MMM & 23 & 34 & 59 & 112 & 87 & 145 & 263 & 498\\
SAM & \textbf{4} & 6 & 9 & 17 & \textbf{9} & 10 & 16 & 23\\
FCM & 610 & 1209 & 2428 & 4832 & 1999 & 3940 & 7913 & 15719\\
PIM & 560 & 1171 & 2401 & 4806 & 1586 & 3406 & 6817 & 13257\\
KM & 128 & 229 & 404 & 757 & 396 & 703 & 1281 & 2400\\
KM-C & 1147 & 2777 & 4395 & 5233 & 3339 & 13294 & 14912 & 22637\\
FKM & 39 & 49 & 78 & 187 & 133 & 158 & 213 & 369\\
FKM-C & 267 & 346 & 420 & 893 & 913 & 1565 & 1285 & 2036\\
SKM & 121 & 207 & 361 & 672 & 380 & 653 & 1173 & 2174\\
\emph{WSM} & 25 & 32 & 57 & 173 & \textbf{9} & 15 & 34 & 136\\
\emph{WSM-C} & 85 & 109 & 182 & 572 & 24 & 34 & 94 & 356\\
\hline
\end{tabular}
}
\end{table}

\begin{table}[ht]
\linespread{1}
\centering
\scriptsize
{
\caption{ \label{tab_perf_rank} Performance rank comparison of the quantization methods}
\begin{tabular}{|c|c|c||c|}
\hline
Method & MSE rank & Time rank & Mean rank\\
\hline
MC & 13.97 & \textbf{1.38} & 7.67\\
WAN & 13.66 & 2.84 & 8.25\\
WU & 8.47 & 3.31 & 5.89\\
NEU & 6.31 & 6.00 & 6.16\\
MMM & 12.31 & 7.63 & 9.97\\
SAM & 10.09 & 2.53 & 6.31\\
FCM & 10.31 & 13.94 & 12.13\\
PIM & 9.81 & 12.94 & 11.38\\
KM & 7.56 & 11.34 & 9.45\\
KM-C & 3.03 & 15.00 & 9.02\\
FKM & 7.91 & 7.75 & 7.83\\
FKM-C & 3.88 & 11.53 & 7.70\\
SKM & 8.06 & 10.25 & 9.16\\
\emph{WSM} & 3.56 & 5.28 & \textbf{4.42}\\
\emph{WSM-C} & \textbf{1.06} & 8.25 & 4.66\\
\hline
\end{tabular}
}
\end{table}

\newpage

\begin{table}[ht]
\linespread{1}
\centering
\scriptsize
{
\caption{ \label{tab_stable_rank} Stability rank comparison of the quantization methods}
\begin{tabular}{|c|c|}
\hline
Method & MSE rank\\
\hline
FCM & 9.36\\
PIM & 9.56\\
KM & 8.31\\
KM-C & 2.84\\
FKM & 8.10\\
FKM-C & 3.41\\
SKM & 7.11\\
\emph{WSM} & 3.92\\
\emph{WSM-C} & \textbf{2.02}\\
\hline
\end{tabular}
}
\end{table}

\end{document}